\documentclass{article}
\usepackage{ijcai23}

\usepackage{times}
\usepackage{soul}
\usepackage{url}
\usepackage[hidelinks]{hyperref}
\usepackage[utf8]{inputenc}
\usepackage[small]{caption}
\usepackage{graphicx}
\usepackage{amsmath}
\usepackage{amsthm}
\usepackage{booktabs}
\usepackage{algorithm}
\usepackage{algorithmic}
\usepackage[switch]{lineno}

\usepackage{epsfig}%
\usepackage{graphicx}
\usepackage{comment}
\usepackage{amsmath,amssymb} 
\usepackage{color}
\usepackage{colortbl}
\usepackage{dirtytalk}%
\usepackage{amssymb}%
\usepackage{mathtools}%
\usepackage{sidecap}
\usepackage[normalem]{ulem}%
\usepackage{multirow}
\usepackage{subfigure}
\usepackage{longtable}
\usepackage{supertabular}
\usepackage{siunitx}
\usepackage{tabu}%
\usepackage{array}
\usepackage{makecell}

\usepackage{amsfonts}
\usepackage{latexsym}
\usepackage[usenames,dvipsnames,svgnames,table]{xcolor}
\usepackage{xspace}
\makeatletter
\DeclareRobustCommand\onedot{\futurelet\@let@token\@onedot}
\def\@onedot{\ifx\@let@token.\else.\null\fi\xspace}
\def\eg{\emph{e.g}\onedot} 

\def\ie{\emph{i.e}\onedot}

\def\etc{\emph{etc}\onedot}

\def\etal{\emph{et al}\onedot}
\makeatother

\def\Vec#1{{\boldsymbol{#1}}}

\definecolor{Gray}{gray}{0.85}
\definecolor{LightCyan}{rgb}{0.88,1,1}

\usepackage{float}
\usepackage{hyperref}

\title{Hyperbolic Geometry in Computer Vision: A Survey}

\author{
Pengfei Fang$^1$
\and
Mehrtash Harandi$^1$
\and
Trung Le$^1$
\and
Dinh Phung$^1$
\affiliations
$^1$Monash University, Australia\\
\emails
\{pengfei.fang, mehrtash.harandi, trunglm, dinh.phung\}@monash.edu
}

\begin{document}

\maketitle

\begin{abstract}
Hyperbolic geometry, a Riemannian manifold endowed with constant sectional negative curvature, has been considered an alternative embedding space in many learning scenarios, \eg, natural language processing, graph learning, \etc, as a result of its intriguing property of encoding the data's hierarchical structure (like irregular graph or tree-likeness data). Recent studies prove that such data hierarchy also exists in the visual dataset, and investigate the successful practice of hyperbolic geometry in the computer vision (CV) regime, ranging from the classical image classification to advanced model adaptation learning. This paper presents the first and most up-to-date literature review of hyperbolic spaces for CV applications. To this end, we first introduce the background of hyperbolic geometry, followed by a comprehensive investigation of algorithms, with geometric prior of hyperbolic space, in the context of visual applications. We also conclude this manuscript and identify possible future directions. 
\end{abstract}

\section{Introduction}\label{sec:intro}

Learning an embedding space to understand the underlying data distribution is a fundamental problem in machine learning whose developed embedding algorithms possibly benefit a diverse set of real-world applications in natural language processing (NLP), computer vision (CV), graph analysis, \etc. The initial embedding algorithms are studied in the familiar Euclidean space (\ie, a vector space with a flat structure enjoying the closed-form formulations for basic operations, \eg, vector addition, inner product, averaging, \etc), which can be considered as a natural generalization of the 3D space. This, in turn, leads to learning a flat embedding space from embedding algorithms, accommodating low-dimensional latent embedding of the raw data.

\begin{figure}[ht]
\centering
\subfigure[An illustration of the graph embedding in the Euclidean space.]{\includegraphics[width=0.54\linewidth]{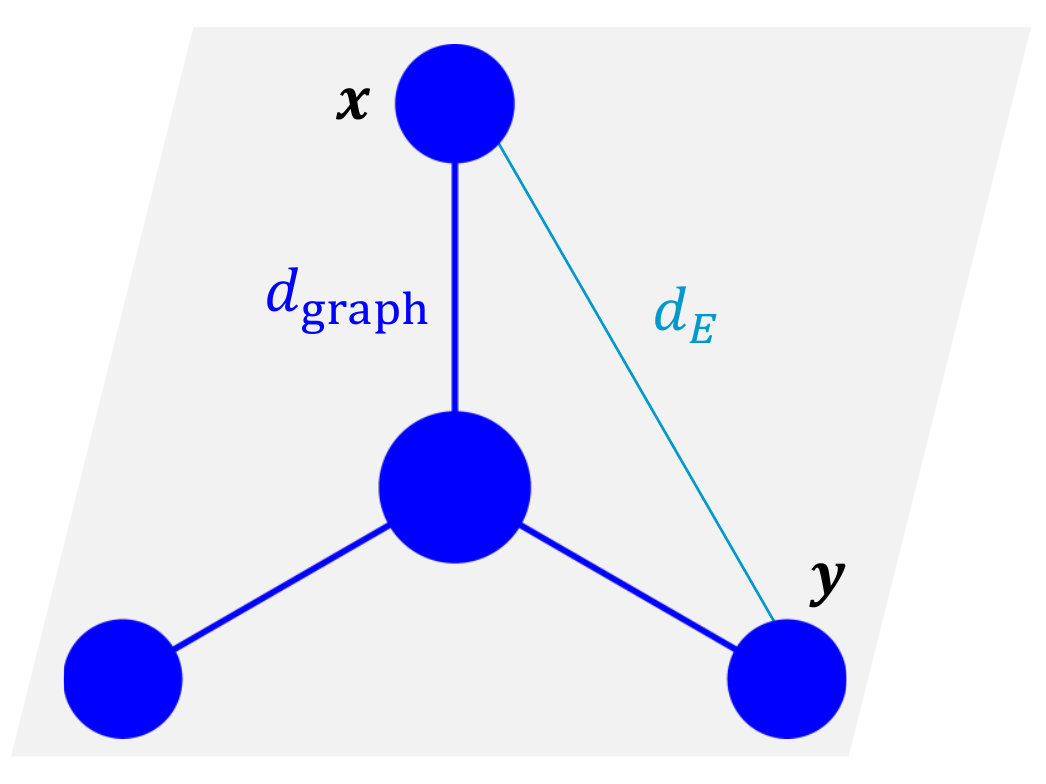}\label{fig:distortion}}%
\hfil
\subfigure[An illustration of 2-D Poincar\'e ball.]{\includegraphics[width=0.4\linewidth]{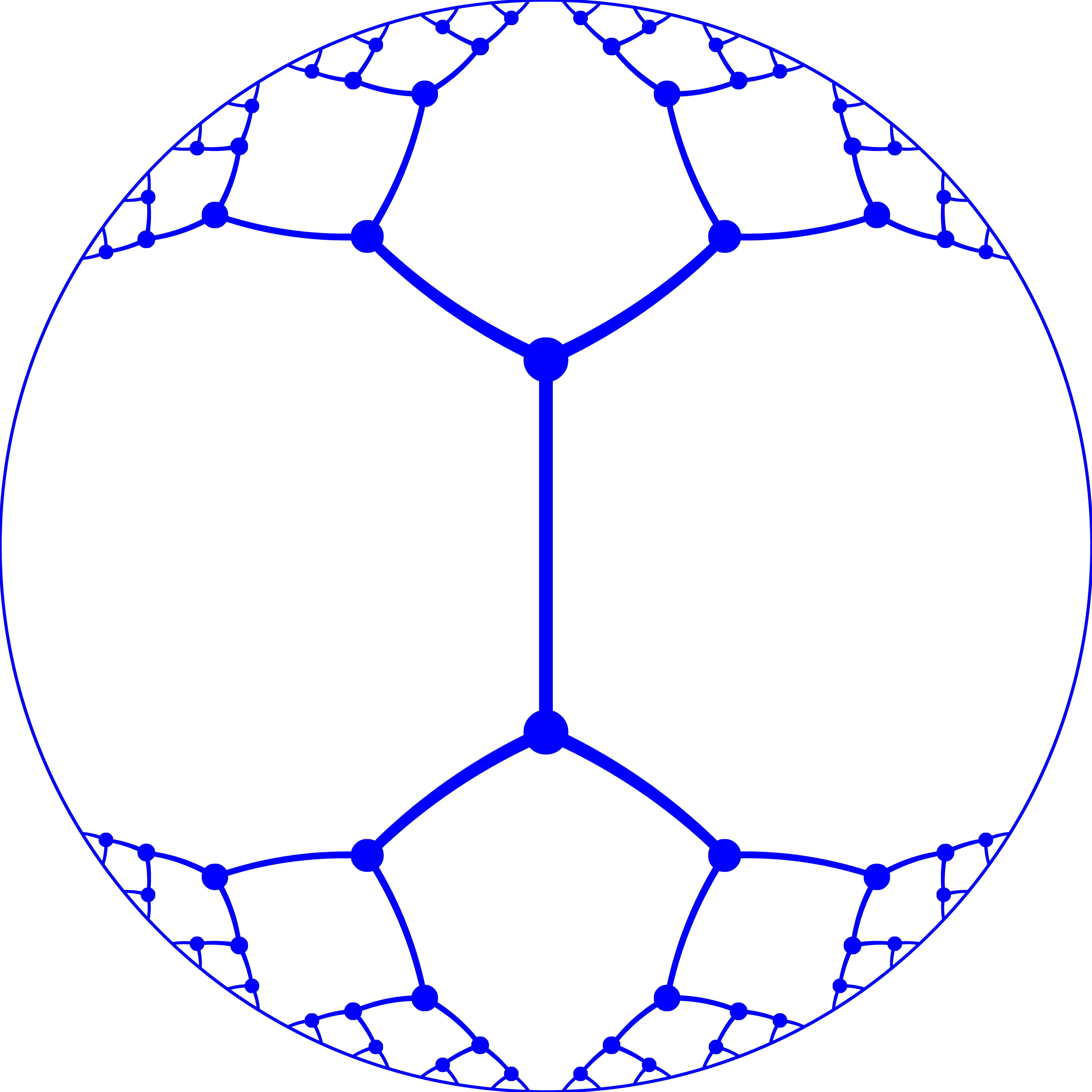}\label{fig:poincare}}%
\hfil
\subfigure[An illustration of the hierarchical structure of cat taxonomy.]{\includegraphics[width=1\linewidth]{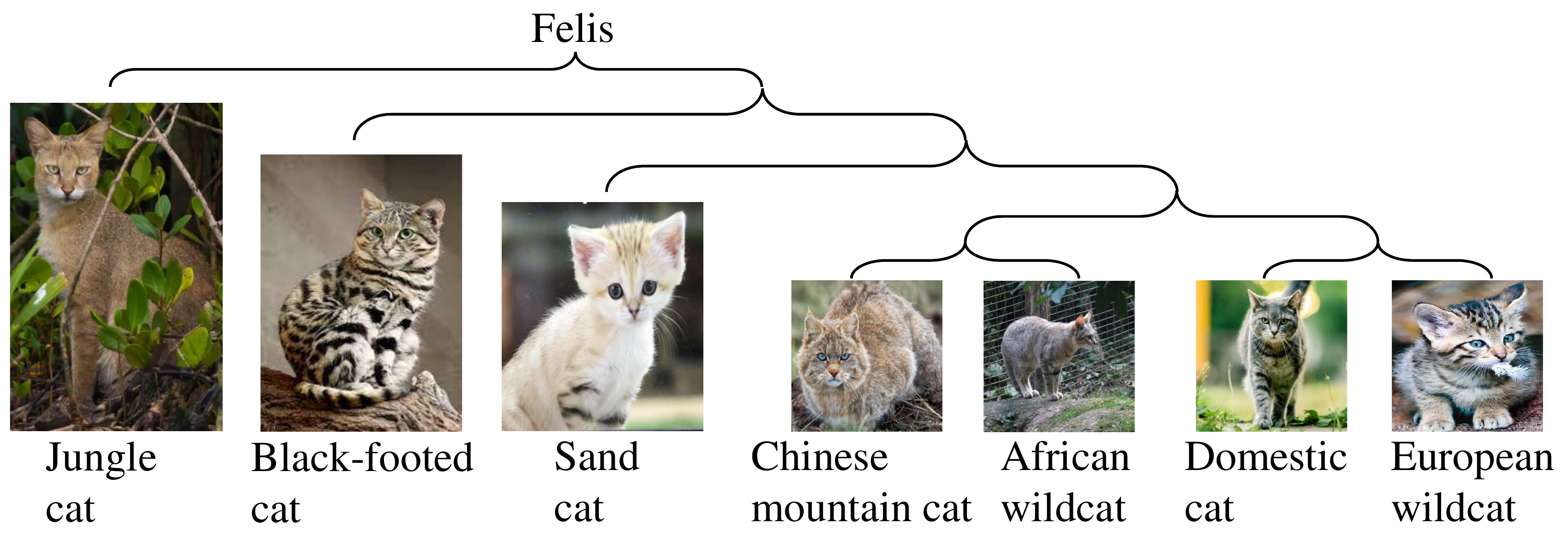}\label{fig:cat}}%
\caption{In (a), for $\Vec{x}$ and $\Vec{y}$, the Euclidean distance and graph distance are $d_{E}(\Vec{x}, \Vec{y}) = \sqrt{3}$ and $d_{\mathrm{graph}}(\Vec{x}, \Vec{y}) = 2$, respectively. The distortion value is $\mathrm{DV} = {d_{\mathrm{graph}}(\Vec{x}, \Vec{y})}/{d_{E}(\Vec{x}, \Vec{y})} = {2}/{\sqrt{3}}$. (b) shows a 2-D Poincar\'e disk that can preserve the data hierarchy. (c) illustrates the hierarchical structure of cat taxonomy, showing the data hierarchy exists in the visual data.}\label{fig:comparison}
\end{figure}

However, despite its simple formulas and wide usage, Euclidean space is not an optimal choice as an embedding space for datasets possessing complicated structures, \eg, hierarchical structure, due to the distortion caused in the Euclidean space. The following toy example is used to support the motivation. For a graph embedded in Euclidean space, as shown in Fig.~\ref{fig:distortion}, the Euclidean distance between $\Vec{x}$ and $\Vec{y}$ can be calculated as $d_{E}(\Vec{x}, \Vec{y}) = \sqrt{3}$. While the graph distance for the nodes $\Vec{x}$ and $\Vec{y}$ is $d_{\mathrm{graph}}(\Vec{x}, \Vec{y}) = 2$. In another word, it causes distortion\footnote{In this context, the distortion value ($\mathrm{DV}$) can be computed as $\mathrm{DV} = {d_{\mathrm{graph}}(\Vec{x}, \Vec{y})}/{d_E(\Vec{x}, \Vec{y})} = 2/\sqrt{3}$.} when embedding such graph data in the Euclidean space, thereby failing to encode the inherent hierarchical structures. The hyperbolic geometry emerges for rescue, being able to accommodate such graph-structured or tree-likeness data (see Fig.~\ref{fig:poincare} of the Poincar\'e ball as the model for hyperbolic geometry).

The hierarchical structure naturally exists in many practical AI applications, such as textual entailment analysis~\cite{Ganea2018_HyperbolicNN_NIPS}, recommendation system~\cite{VinhTran2020HyperMLAB}, social network analysis~\cite{Chami2019_HyperbolicGCNN_NIPS}, \etc. In such applications, the data format intuitively matches the hierarchical structure, and the hyperbolic geometry is proven to work particularly well for applications processing with such tree-likeness data. Its rich representation power, of late, is also successfully applied to the visual embeddings \cite{Yoshihiro2019_WrappedND,Khrulkov_2020_CVPR}, in conjunction with the observation that the hierarchical structure also exists in the vision dataset (see Fig.~\ref{fig:cat}).

Recently, Peng \etal contributed a comprehensive literature review for the hyperbolic deep neural networks (HDNN), which summarizes the neural components in the construction of HDNN, the deep learning algorithms developed for the hyperbolic geometry, and applications~\cite{Wei2022_hyperbolic_PAMI}. However, it does not include an up-to-date literature review for vision applications. Specifically, the CV research empowering intelligent machines the capability of sensing and understanding the visual world, is paramount in modern AI technologies. Additionally, due to the tensor format of the visual data, it also requires to develop algorithms tailored for the visual tasks. This motivates us to summarize the advances of hyperbolic geometry in the CV field. As a complementary of the work~\cite{Wei2022_hyperbolic_PAMI}, this manuscript for the first time presents the progress of approaches developed on top of the hyperbolic geometry in visual applications. We believe our survey paper will provide a high-level picture of the development on hyperbolic algorithms for the CV applications, and encourages to develop more efficient and effective methods.

This survey also aims to answer the question that \say{\textit{along with the superior performance, what else we can conclude from the literature about the properties of hyperbolic geometry that benefit CV applications?}}. Three important observations are drawn from the literature. 1) The hyperbolic embeddings bring a significant performance gain to the Euclidean embedding in a lower dimensional space~\cite{Yoshihiro2019_WrappedND,Khrulkov_2020_CVPR}. One conjecture is that the hyperbolic space can embed hierarchies with low distortion into a low dimensional space, while Euclidean space cannot. This observation also verifies that the hierarchical structure exists in image datasets. 2) The hierarchical nature of hyperbolic geometry makes it possible to measure the confidence of the sample via its location~\cite{Khrulkov_2020_CVPR,Atigh_2022_CVPR}. In the Poincar\'e ball, the ambiguous, unclear, or uncertain samples are embedded near the origin, while the confident/clear samples lie near the boundary. Also, the location can also indicate the relation of samples, known as the parent-children relation of nodes~\cite{suris2021hyperfuture}. 3) The hyperbolic representations can outperform the Euclidean representations in the tasks of low data regime \cite{Khrulkov_2020_CVPR,Liu_2020_CVPR,Fang_2021_ICCV}, like few-shot learning, zero-shot learning \etc, showing its better generalization property. Those observations encourage future developments in hyperbolic geometry.

The rest of this paper is organized as follows: \textsection~\ref{sec:hyperbolic} introduces the background of the hyperbolic geometry, including the definition and its modeling, \ie, Poincar\'e ball, in the CV domain. A complete review of the recent advances of hyperbolic algorithms for CV applications is then presented in \textsection~\ref{sec:hyperbolic_alg}. Finally, \textsection~\ref{sec:conclusion} concludes this paper and discusses possible future directions.

\section{Hyperbolic Geometry and Poincar\'e Ball}\label{sec:hyperbolic}

In this section, we will introduce background knowledge of the hyperbolic geometry and its modeling, \ie, Poincar\'e ball. 

The $n$-dimensional hyperbolic space $\mathbb{H}^n$ is defined as a $n$-dimensional Riemannian manifold with a constant negative curvature. Five isometric models, including the Poincar\'e ball model, the Poincar\'e half-space model, the Lorentz model, the Klein model, and the Hemisphere model, are employed to model the hyperbolic space~\cite{hyperbolic_1997}. In this survey, we are particularly interested in the Poincar\'e ball model, in the sense that most works leverage the Poincar\'e ball model to work with the hyperbolic geometry in the CV field. The rest of this section will explain the Poincar\'e ball.

The Poincar\'e ball, as a model of $n$-dimensional hyperbolic geometry, can be described via a manifold and a $n$-dimensional Riemannian metric, denoted by $(\mathbb{D}^n, g^{\mathbb{D}})$. The manifold $\mathbb{D}^n$ is defined as $\mathbb{D}^n = \{\Vec{x} \in \mathbb{R}^n: \|\Vec{x}\|^2 < 1 \}$, and the Riemannian metric $g^{\mathbb{D}}$ is given by $g^{\mathbb{D}}(\Vec{x}) = \lambda^2(\Vec{x})g^{E}$, where $\lambda(\Vec{x}) = \frac{2}{1-\|\Vec{x}\|^2}$ is the conformal factor at $\Vec{x}$ and $g^{E}$ is the Euclidean metric.

The shortest path, connecting a pair of points along the manifold, is termed {geodesic}\footnote{In the Euclidean space, the geodesic is a straight line connecting two points.}. The length of the geodesic is called {geodesic distance}. Specifically, for any two points $\Vec{x}_i,~\Vec{x}_j \in \mathbb{D}^n$, the geodesic distance between $\Vec{x}_i$ and $\Vec{x}_j$ is:

\begin{equation}
d_{\mathbb{D}}(\Vec{x}_i, \Vec{x}_j) = \mathrm{cosh}^{-1}\big( 1+ 2\frac{\|\Vec{x}_i - \Vec{x}_j\|^2}{(1-\|\Vec{x}_i\|^2)(1-\|\Vec{x}_j\|^2)} \big),   
\end{equation}
where $\mathrm{cosh}^{-1}(\cdot)$ is the inverse hyperbolic cosine function. 


In the Euclidean space, the vector operations inherited from the flat vector space, \eg, vector addition or multiplication, are integrated components of the neural architecture and learning procedure. However, using such standard operations in the Poincar\'e ball is not without difficulties as a result of its curved geometry. To facilitate vector operations in the Poincar\'e ball, the M\"obius gyrovector space may come in handy. That said, one can resort the algebraic formalism, provided by the  M\"obius gyrovector space, to work with the Poincar\'e ball.

In the M\"obius gyrovector space, one can further use an extra hyperparameter, \eg, curvature $-c$ for $c>0$, to define the Poincar\'e ball, as $\mathbb{D}_c^n = \{\Vec{x} \in \mathbb{R}^n: c\| \Vec{x} \|^2 < 1 \}$. Then its conformal factor takes the form of $\lambda_c(\Vec{x}) = \frac{2}{1 - c\|\Vec{x}\|^2}$.

To enable vector operations, we first introduce the {M\"obius addition}. The M\"obius addition for $\Vec{x}_i, \Vec{x}_j \in \mathbb{D}_c^n$ is:

\begin{equation}
\Vec{x}_i \oplus_c \Vec{x}_j = \frac{(1 + 2c\langle \Vec{x}_i, \Vec{x}_j\rangle + c\|\Vec{x}_j\|^2)\Vec{x}_i + (1 - c\|\Vec{x}_i\|^2)\Vec{x}_j}{1 + 2c\langle\Vec{x}_i, \Vec{x}_j\rangle + c^2\|\Vec{x}_i\|^2\|\Vec{x}_j\|^2}.
\end{equation}

The induced geodesic distance for $\Vec{x}_i$ and $\Vec{x}_j$ on $\mathbb{D}_c^n$ is formulated as:

\begin{equation}\label{eq:geodis}
d_{\mathbb{D}_c}(\Vec{x}_i, \Vec{x}_j) = \frac{2}{\sqrt{c}}\mathrm{tanh}^{-1}(\sqrt{c}\|-\Vec{x}_i \oplus_c \Vec{x}_j\|),
\end{equation}
where $\tanh^{-1}(\cdot)$ is the inverse hyperbolic tangent function.

\begin{figure}[ht] 
\centering
\includegraphics[width=0.84\linewidth]{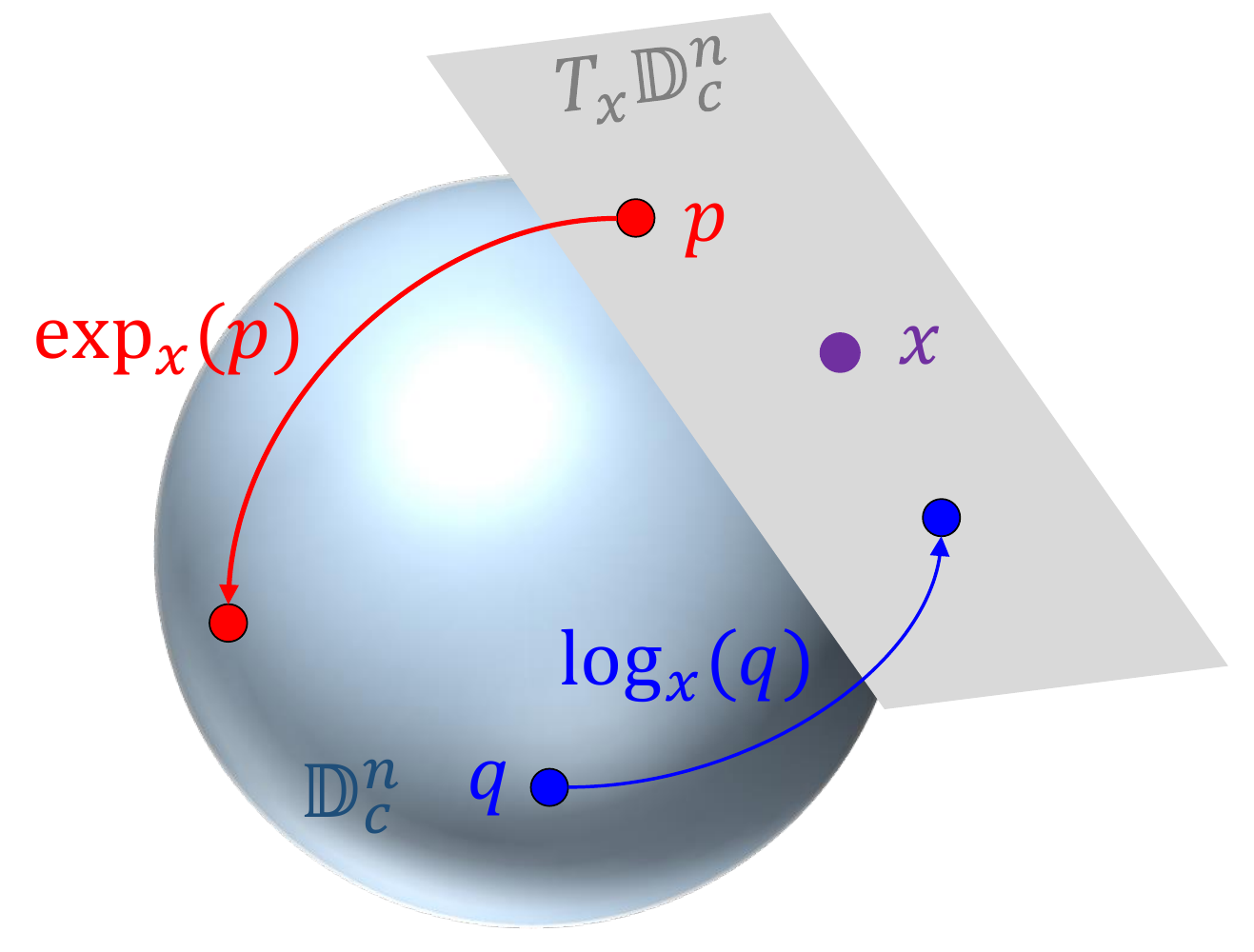}
\caption{An illustration of the {Poincar\'e ball ($\mathbb{D}_c^n$)}, {tangent plane ($T_{\Vec{x}}\mathbb{D}_c^n$)}, and the associated mappings between them, \ie, {exponential mapping ($\exp_{\Vec{x}}(\cdot)$)} and {logarithm mapping ($\log_{\Vec{x}}(\cdot)$)}.} \label{fig:manifold_mapping}
\end{figure}

For a point $\Vec{x} \in \mathbb{D}^n_c$, the tangent space of $\mathbb{D}_c^n$ at $\Vec{x}$, denoted by $T_{\Vec{x}}\mathbb{D}_c^n$, is a $n$-dimensional vector space which is a first order approximation of $\mathbb{D}_c^n$ at $\Vec{x}$. Fig.~\ref{fig:manifold_mapping} shows $\mathbb{D}_c^n$ and $T_{\Vec{x}}\mathbb{D}_c^n$. To work with the hyperbolic space freely, a bijective mapping between $\mathbb{D}_c^n$ and $T_{\Vec{x}}\mathbb{D}_c^n$ is required. The {{exponential map}} $\exp_{\Vec{x}}(\cdot) : T_{\Vec{x}}\mathbb{D}_c^n \to \mathbb{D}_c^n$, provides a way to project a point $\Vec{p} \in T_{\Vec{x}}\mathbb{D}^n_c$ to $\mathbb{D}^n_c$, as follows:

\begin{equation}\label{eq:exp}
\begin{split}
\exp_{\Vec{x}}(\Vec{p}) = \Vec{x}\oplus_c\big(\tanh(\sqrt{c}\frac{\lambda_c({\Vec{x}}) \cdot \|\Vec{p}\|}{2}) \frac{\Vec{p}}{\sqrt{c}\|\Vec{p}\|}\big),
\end{split}
\end{equation}
where $\tanh(\cdot)$ is the hyperbolic tangent function.


The inverse process of exponential map is termed {logarithm map}. The logarithm map $\log_{\Vec{x}}(\cdot) : \mathbb{D}_c^n \to T_{\Vec{x}}\mathbb{D}_c^n$ projects a point $\Vec{q} \in \mathbb{D}^n_c$, to the tangent plane of $\Vec{x}$ (\ie, $T_{\Vec{x}}\mathbb{D}^n_c$), as:

\begin{equation}\label{eq:log}
\begin{split}
\log_{\Vec{x}}(\Vec{q}) = \frac{2}{\sqrt{c} \lambda_c({\Vec{x}})} \tanh^{-1}(\sqrt{c} \|-\Vec{x}\oplus_c  \Vec{q}\|) \frac{-\Vec{x}\oplus_c  \Vec{q}}{\|-\Vec{x}\oplus_c  \Vec{q}\|}.
\end{split}
\end{equation}

Fig.~\ref{fig:manifold_mapping} demonstrates the associated mappings between the manifold and the tangent plane. Of note, it also holds that $\log_{\Vec{x}}\big(\exp_{\Vec{x}}(\Vec{p})\big) = \Vec{p}$.

On top of the basic mappings in Eq.~\eqref{eq:exp} and \eqref{eq:log}, many useful transformations, that realize essential components in the deep neural networks, are also defined. The operators include fully-connected (FC), multi-class logistic regression (MLR), activation function, \etc Please refer to~\cite{Wei2022_hyperbolic_PAMI} for a comprehensive review of the hyperbolic neural operations.

It's also worth noting how to integrate the geometry constraint in practice. In doing so, a transformation layer is used to realize a project function that maps data embeddings from Euclidean space to hyperbolic space as: $\Vec{z} = \mathrm{Proj}(\Vec{x})$ for  $\Vec{x} \in \mathbb{R}^n$ and $\Vec{z} \in \mathbb{D}^n_c$. Table~\ref{table:practice} illustrates the instantiation for the project function. A common practice in ~\cite{Khrulkov_2020_CVPR,Hyperbolic3D2020_ECCV} instantiates $\mathrm{Proj}(\cdot)$ by $\exp_{\Vec{0}}(\cdot)$, followed by a constraint as follows:

\begin{equation}\label{eq:clip_norm}
\Vec{z} = \Gamma(\Vec{p}) = 
\begin{cases}
\Vec{p} & \text{if}~~\|\Vec{p} \| < \frac{1}{\sqrt{c}} \\
\frac{1-\xi}{\sqrt{c}} \frac{\Vec{p}}{\|\Vec{p}\|} & \text{else},
\end{cases}                
\end{equation}
where $\Vec{p} = \exp_{\Vec{0}}(\Vec{x})$, and $\xi$ is a tiny value to ensure the numerical stability. 

\cite{Fang_2021_ICCV} argue that the above projection can not fully utilize the hyperbolic space, as every vector is flattened at the identity and only the vectors close to the origin can approximate the structure in hyperbolic spaces. It brings another practice only applying the constraint in Eq.~\eqref{eq:clip_norm} to the output of the feature extractor, such that the network can optimize the network to encode the input data to the hyperbolic space directly~\cite{Fang_2021_ICCV,ma2022adaptive}.

In the existing deep hyperbolic networks, a hybrid architecture where a neural network extracts feature embedding of the input data in the Euclidean space, followed by a transformation layer to obtain the hyperbolic embedding, is adopted. The most recent work~\cite{Guo_2022_CVPR_Clipped} theoretically analyzes that it is easy to cause a vanishing gradient during back-propagation in the hybrid architecture, restricting the network's applicability. Guo \etal alleviate this issue by simply clipping the Euclidean embedding, formulated as:
\begin{equation}\label{eq:feature_clip}
\Vec{p} = \Phi(\Vec{x}) = \min\{1, \frac{r}{\|\Vec{x}\|}\}\cdot \Vec{x},
\end{equation}
where $r$ is a hyper-parameter. Then the clipped embedding is further projected to hyperbolic space via exponential mapping as $\Vec{z} = \exp_{\Vec{0}}(\Vec{p}) =  \exp_{\Vec{0}}\big( \Phi(\Vec{x}) \big)$. This process can be understood as first bounding the Euclidean embedding within an $r$-radius sphere, then projecting to the hyperbolic space. This ensures all embeddings are located in an open ball of the radius of $2r$~\cite{Guo_2022_CVPR_Clipped}.

\begin{table}[!ht]
\begin{center}
\scalebox{0.96}
{
\begin{tabular}{lc}

\hline
\multirow{2}{*}{Ref.} & \multirow{2}{*}{$\Vec{z} = \mathrm{Proj}(\Vec{\Vec{x}})$; $\Vec{x} \in \mathbb{R}^n$, $\Vec{z} \in \mathbb{D}^n_c$} \\
&\\
\hline

\cite{Khrulkov_2020_CVPR} & $\Vec{z} = \Gamma\big( \exp_{\Vec{0}}(\Vec{x}) \big)$\\

\cite{Fang_2021_ICCV} & $\Vec{z} = \Gamma(\Vec{x})$ \\

\cite{Guo_2022_CVPR_Clipped} &  $\Vec{z} =  \exp_{\Vec{0}}\big( \Phi(\Vec{x}) \big)$\\

\hline

\end{tabular}
}
\end{center}
\caption{Summary of the instantiation for the project function (\ie, $\mathrm{Proj}(\cdot)$).}
\label{table:practice}
\end{table}

\section{Hyperbolic Algorithms in Computer Vision}\label{sec:hyperbolic_alg}

Building intelligent machines that can understand the visual concept like humans is an essential branch in AI, due to its significant industrial potential, like autonomous driving, augmented reality, medical image analysis \etc, as well as the academic importance in terms of creating a discriminative latent embedding space that can reveal the underlying pattern of the raw visual data (\eg, image, video, point cloud, \etc). The Euclidean space with flat structures has been the default option of the embedding space for the visual data. Recent years, the hyperbolic geometry has gained increasing interest as an alternative to the embedding space, and the CV community has benefited from the hyperbolic space since 2019, with one of the earliest works being~\cite{Emile2019_HVAE}. These works extensively investigate the advantages of hyperbolic geometry in the CV field (see Table~\ref{table:papers}) and we group those works into three categories, namely unsupervised learning, supervised learning, and model adaptation learning. In the following, an up-to-date review will be made of the advanced approaches.

\subsection{Unsupervised Learning}

In the visual embedding learning, hyperbolic geometry starts to show its advantages in unsupervised learning, \eg, clustering, and variational auto-encoder. These early works attempt to address the difficulties of realizing the algorithms that are aware of the geometry prior.

Clustering, which can automatically discover the data grouping without annotations, is a popular unsupervised learning paradigm~\cite{Dongkuan_clustering}. Posing the prior of that data groups are represented as a dendrogram, hierarchical clustering (HC), a.k.a, connectivity-based clustering, comes into mind~\cite{Akshay_HC_ICML}. In general, the HC approaches establish a hierarchy relationship over clusters, such that the clustering result forms a multi-layered tree, with leaves corresponding to samples and internal nodes corresponding to clusters. Despite the significant progress that has been made to improve the clustering quality~\cite{Flonn_hierarchical_clustering}, such methods are restricted by the discrete nature of trees, including the scalable stochastic optimization or expensive computation cost. Having those issues in mind, the hyperbolic geometry, which can be understood as a continuous representation of a tree, is used as an alternative space for HC. This motivates the community to study the HC in hyperbolic space, called hyperbolic hierarchical clustering (HHC)~\cite{Nicholas_2019_KDD,NEURIPS2020_ac10ec1a_HypHC,CMHHC_IJCAI2022}.

In~\cite{Nicholas_2019_KDD}, gradient-based hyperbolic hierarchical clustering (gHHC) searches a continuous tree space by considering the child-parent relationships, \eg, relative position and norm of node embeddings, optimized by stochastic gradient descent (SGD) manner. Specifically, the norm of node embeddings can indicate the depth of the tree. The child nodes are required to have a larger norm than that of their parent nodes, such that root nodes are located near the origin and the leaves are distributed near the boundary of the ball. Such geometric prior benefits the existing HC loss to optimize the embeddings with the SGD inference procedure. HYPHC in \cite{NEURIPS2020_ac10ec1a_HypHC}, of later, formulates a continuous relaxation of Dasgupta's discrete optimization~\cite{Dasgupta_HC}. That is, the embedding of the leaf node in the hyperbolic space can be optimized in a continuous search space and an auxiliary decoder recovers a discrete tree for continuous representation. \cite{CMHHC_IJCAI2022} improve HYPHC by first learning meaningful instance embeddings in a proper metric space, benefiting the final HC performance.

\cite{Yan_2021_CVPR} adopt the HC in a real-world application of deep metric learning (DML), and dub the algorithm as unsupervised hyperbolic metric learning (UHML). UHML first follows the common practice that projects the Euclidean embeddings to the hyperbolic space, and leverages the HC to produce pseudo-hierarchical labels. To benefit from the pseudo-hierarchical labels, a log-ratio loss is formulated to push the log of distance ratio (\eg, $\log(\|\Vec{z}_i - \Vec{z}_j\| / \|\Vec{z}_i - \Vec{z}_l\|)$) to the log of similarity difference (\eg, $\log(\Omega^{s_{ij} - s_{il}})$) for a triplet (\eg, $(\Vec{z}_i, \Vec{z}_j, \Vec{z}_l)$), thereby learning a proper hierarchical metric space.

The variational auto-encoder (VAE) is another popular generative model in unsupervised learning, and it learns the data distribution in a low-dimensional latent space in a probabilistic inference manner~\cite{Kingma2014_VAE,pmlr-v32-rezende14}. That is, instead of building an encoder that produces a static latent embedding, the VAE formulates the encoder to regress a probability distribution per latent attribute. Many studies extensively explore its potential in embedding learning and generation capacity~\cite{SVAE18_UAI,NEURIPS2018_WVI,Yoshihiro2019_WrappedND,Emile2019_HVAE}, and most of improvements are developed in the Euclidean space. Considering the benefits of hyperbolic space, one can further yield such a probabilistic model in the hyperbolic space when the hierarchical structure arises in the dataset~\cite{Yoshihiro2019_WrappedND,Emile2019_HVAE}. To integrate the variational inference of VAEs in hyperbolic spaces, it is required to define a probability distribution. \eg, Gaussian distribution, in the hyperbolic space as a prior.

\cite{Yoshihiro2019_WrappedND} bridge this gap by developing a pseudo-hyperbolic Gaussian distribution, which leverages the operations of parallel transport and exponential map in the hyperbolic space to realize the reparameterization process in VAE~\cite{Kingma2014_VAE}. Specifically, having the pseudo-hyperbolic Gaussian distribution $\mathcal{H}(\Vec{\mu}, \Sigma)$ at hand, one first sample a vector in a vanilla Gaussian distribution $\mathcal{N}(\Vec{0}, \sigma)$ and interpret the vector as a point in the tangent plane of the origin. Then the parallel transport is employed to deliver the vector to the tangent plane of $\Vec{\mu}$, and the exponential map is used to project the delivered vector to the hyperbolic space. This work only identifies the Gaussian distribution in the Lorentz model and does not pose the geometry constraints in the network architecture. Along with deriving two closed-form of Gaussian distribution in the Poincar\'e ball (\eg, Riemannian normal and wrapped normal), the work in \cite{Emile2019_HVAE} develops a gyroplane layer, such that both the encoder and decoder take into account the geometry constraint in the latent space. Briefly, it modifies the output of the encoder and the input of the decoder to work with the gyroplane layer. The encoder produces a Fr\'echet mean and distortion to describe a Gaussian distribution in the hyperbolic space, and the gyroplane layer, as the first layer of the decoder, realizes an affine transformation for the hyperbolic data. Both two approaches show superior performance to the Euclidean counterpart on the digit reconstruction task, especially in a lower-dimension latent space. However, they lack the evaluation of large-scale datasets on both methods. In addition, the inherent drawback of the VAE paradigm, \ie, posterior collapse, also calls for the effort to develop more stable generative methods, \eg, VQ-VAE~\cite{NIPS2017_VQVAE} or diffusion model~\cite{NIPS2020_DenoiseDiffusion}, in the hyperbolic space.

Those two popular unsupervised learning paradigms, endowed with the hyperbolic geometry, also demonstrate their superiority in unsupervised image segmentation task~\cite{NEURIPS2021_291d43c6,weng2021unsupervised_cvpr}. The idea of hyperbolic VAE is extended in~\cite{NEURIPS2021_291d43c6} to solve the challenging semantic segmentation task for biomedical images, a 3D voxel-grid format of data. On top of the model in~\cite{Emile2019_HVAE}, \cite{NEURIPS2021_291d43c6} develop a 3D hyperbolic VAE and a corresponding 3D gyroplane layer to process the 3D patches, sampled from the biomedical image. A regularization term, hierarchical triplet loss, is also proposed to explicitly encode the hierarchical structure. In a whole of input biomedical image, any sampled 3D patch is used as an anchor, while the positive is a sub-patch of the anchor, and the negative is another 3D patch of the whole biomedical image. The rich representation power of hyperbolic geometry also promotes the discovery of novel objects for instance segmentation, studied in~\cite{weng2021unsupervised_cvpr}. For an image with unseen objects, it first uses a pre-trained mask proposal network to produce a class-agnostic mask proposal, followed by learning the embedding in the hyperbolic space, optimized by triplet loss with sampled mask triplets. Having a trained Poincar\'e ball at hand, unsupervised hyperbolic clustering is employed to identify the distinct objects, realizing the instant segmentation for novel samples.

\subsection{Supervised Learning}\label{sec:SL}

Classification or recognition task for fixed-set is the most fundamental problem in the supervised learning paradigm. It optimizes a classifier to predict the label of the input data. Its paradigm in the existing CV work adopts a hybrid fashion to realize the classifier in hyperbolic spaces. A feature extractor first produces a Euclidean embedding, followed by a constraint made to project the Euclidean embedding to the hyperbolic space. Then a hyperbolic counterpart of classifier, known as MLR, is stacked for the classification task.

This hybrid paradigm is explored to address challenging CV tasks extensively~\cite{Khrulkov_2020_CVPR,Hyperbolic3D2020_ECCV,Ola2022Fisheye}. The pioneer work~\cite{Khrulkov_2020_CVPR} first shows the hierarchical structure indeed exists in the image dataset and successfully studies the possibility of employing hyperbolic operations in the classification task. Later, Chen \etal propose the Hyperbolic Embedded Attentive representation (HEAR) model, which explicitly builds the hierarchy in the 3D data, and encodes the hierarchical 3D representations~\cite{Hyperbolic3D2020_ECCV}. HEAR receives multi-view of images as the input for a 3D object, and develops a hybrid attention (HA) mechanism and multi-granular view pooling (MVP) layer, to encode coarse-to-fine hierarchical embeddings per 3D object. To properly enable the hierarchical relations in the embedding space, such embeddings are further projected to the Poincar\'e ball.

In contrast to image-level classification task, the image segmentation can be modeled as pixel-level classification task. The image segmentation leverages the hyperbolic space's property of automatically discovering uncertainties to identify the boundary of the objects~\cite{Atigh_2022_CVPR}. Once a segmentation network is trained with each pixel represented in the Poincar\'e ball, the uncertain pixels, measured by the distance to the origin, align to the semantic boundary of objects. This property enables to predict the semantic boundary for free, leading to produce image segmentation results fast.

Considering the connection between hyperbolic space and the imaging principle of the ultra-wide field-of-view (FoV) image, \cite{Ola2022Fisheye} propose a hyperbolic deformable kernel (HDK) learning strategy for FoV image recognition/segmentation task. HDK also involves a hybrid architecture that learns the deformable kernel, the positions within the kernel window, in the hyperbolic space, and applies the learned deformable kernels to the feature map in the Euclidean space. Its superiority is verified in FoV image segmentation across different distortion level.

Action recognition is also an important problem in CV applications and the hyperbolic space shows its potential in various settings of action recognition tasks. In the skeleton-based action recognition study, the topology structure of the skeleton can be naturally modeled as a graph, and the graph convolutional networks (GCNs) are particularly good at encoding such irregular data. That said, the skeleton-based action recognition can benefit from the hyperbolic space, in the sense that the hyperbolic space can improve the representation power of GCNs. This idea is studied in \cite{Peng2020_ACMMM}. Peng \etal, propose a spatial-temporal GCN (ST-GCN) in the Poincar\'e ball. A mix-dimension method, inspired from~\cite{Yu2018SlimmableNN}, is also proposed in the Poincar\'e ball to identify an optimal dimension of the graph representation.

In addition to the data format, the concept of action hierarchy also draws attention to modeling the action in the hyperbolic space. Considering the diagram in~\cite{Long_2020_CVPR} as an example, the relation of $\{\underline{Taichi}, \underline{Karate}, \underline{Kickbox}\} \to \big[\{\underline{Martial}\}, \{Ball~games\}\big] \to \{\underline{Sport}\}$ constructs the action hierarchy for $\underline{Action}$. Also, $\{Taichi\} \to \{ Matrial \}$ and $ \{ Matrial \} \to \{ Sport \}$ refer to hyponym-hypernym relation.

Such relation of action hierarchy inspires more action recognition tasks, \eg, action search or action prediction, to benefit from hyperbolic space~\cite{Long_2020_CVPR,suris2021hyperfuture}. Long \etal propose a hyperbolic action network that creates a Poincar\'e ball accommodating both the action text and action videos~\cite{Long_2020_CVPR}. It firsts encodes both the text embedding and the action hierarchy, obtaining the action prototypes. Then another neural network is optimized to encode the action videos to a shared Poincar\'e ball, supervised by the action prototypes. The hierarchical nature of the hyperbolic space also helps the action prediction task. Given the fact that the norm of the Poincar\'e embedding can indicate the confidence of the model, the hyperbolic space enables the network to predict either the concrete future action when the network is confident or a higher-level abstraction (\eg, hypernym) if the network is not confident~\cite{suris2021hyperfuture}.

Having the observation that the medical dataset has underlying taxonomy in the disease label space, \cite{Yu2022_MICCAI} encode the class hierarchy for the label text, known as class prototypes, and leverage the class prototypes to supervise the medical image embedding learning, such that the hierarchical structure of the images can be encoded in the latent embedding space. This method has shown its superiority for the skin lesion recognition task.

\subsection{Model Adaptation Learning} 

The rapid progress in the CV domain is driven by learning algorithms that rely on large neural networks and a huge collection of data. Nevertheless, either training a large model or collecting immensely complete data at once for learning is not feasible in real practice. As a result, it rises an idea of open-ended learning for intelligent machines that learn to modify their behaviors for new purposes (\eg, recognizing unknown objects, knowledge transfer or life-long learning) via model adaptation learning algorithms~\cite{ProtoNet_NIPS,KD_hinton,li2016learning}. Such algorithms include learning from a low data regime (\eg, few-shot learning or metric learning), learning to align cross-modality data (\eg, zero-shot learning), learning from teacher models (\eg, knowledge distillation), \etc Such algorithms require models to learn a generalizable or discriminative latent space, which is dominated by Euclidean and spherical geometries in their initial developments.

In the most recent years, some studies start to explore the benefits of hyperbolic space to the model adaptation algorithms~\cite{Khrulkov_2020_CVPR,Liu_2020_CVPR,Ermolov_2022_CVPR}.  \cite{Khrulkov_2020_CVPR} analyze the data hierarchy in the visual dataset, and first demonstrate the potential of hyperbolic space being an alternative embedding space for the visual data. The hyperbolic neural network layers, proposed in \cite{Ganea2018_HyperbolicNN_NIPS}, are successfully employed in the vision tasks, and benefit the few-shot learning (FSL)~\cite{ProtoNet_NIPS} and person re-identification (ReID)~\cite{LiWei2014DeepReID} tasks. Of note, in~\cite{Khrulkov_2020_CVPR}, the prototype of the support set is realized by the Einstein midpoint, instead of the Fr{\'e}chet mean, as the averaging in the hyperbolic space, thereby reducing the computational cost of the network.

Another concurrent work realizes the embedding alignment of visual and semantic features in the Poincar\'e ball~\cite{Liu_2020_CVPR}. This task is known as zero-shot learning (ZSL)~\cite{ALE_Akata_PAMI}, which aims to identify unseen instances via associated attributes. The pipeline first encodes both visual feature and semantic feature in Poincar\'e balls, and then aligns embedding spaces (\eg, two Poincar\'e balls) via minimizing a modified ranking loss~\cite{Facenet_CVPR15}, which pulls the visual embedding and its semantic embedding together, while pushes different semantic embeddings away.

Since then, researchers investigate more possibilities to improve model adaptation approaches in hyperbolic space. Positioning the label embedding of each class at the boundary of the Poincar\'e ball as an ideal prototype, \cite{NEURIPS2021_01259a0c} propose a hyperbolic Busemann learning scheme that jointly steers each sample closer to its class prototypes and forces the ambiguous samples to the origin. In contrast to the work in~\cite{Khrulkov_2020_CVPR}, the hyperbolic knowledge transfer (HyperKT) framework explicitly encodes the class hierarchy for FSL~\cite{HyperKT_IJCAI2022}. HyperKT poses the constraint of class hierarchy to a neural network that encodes semantic embeddings, and acts the semantic embeddings as class prototypes. As auxiliary supervision, such class prototypes transfer the class hierarchy knowledge to the image embedding network, enabling the class hierarchy structure in the image embedding space.

Considering the problem formulation of FSL, one principle way to improve the generalization is to establish a distinctive metric space for new samples. \cite{ma2022adaptive} address this by learning a hyperbolic metric that adaptatively characterizes the point to set (P2S) distance in the Poincar\'e ball. The P2S is formulated as a weighted sum of the point-to-point distance in the tangent space. The weights are learned with a context-aware mechanism, thereby making the metric local optimal. Instead of learning embeddings in a fixed Poincar\'e ball, \cite{Gao_2021_ICCV} generate adaptative geometrical structures for different FSL tasks in the meta-learning training phase\footnote{The FSL is formulated as an episode training manner, and each episode represents an $N$-way $K$-shot classification problem. We term each episode as a task in this context.}. This work has observed that hyperbolic spaces with different curvatures benefit different tasks. This motivates to leverage of a neural network to learn task-dependent curvatures, leading to an optimal geometric structure for each task.

\cite{Fang_2021_ICCV} study positive definite (PD) kernels in hyperbolic spaces, enabling the hyperbolic space to enjoy the rich representation capacity of the kernel machine. This work first reveals a property of the \say{curve length equivalence} theorem between the Poincar\'e ball and the tangent space. This theorem identifies a bijective function, that paves the way to define PD kernels in the Poincar\'e ball. The proposed kernels include hyperbolic tangent kernel, hyperbolic RBF kernel, hyperbolic Laplace kernel \etc, and benefit some model adaptation tasks, \eg, FSL, ZSL, person ReID, and knowledge distillation~\cite{KD_hinton}. However, tuning kernels for each task requires more effort, and how to develop automated kernel learning methods still remains an open problem in this regime.

Towards a general metric learning setting, \cite{Ermolov_2022_CVPR} develop hyperbolic Vision Transformers (Hyp-ViT) that combine the best of two worlds (\ie, hyperbolic geometry and Vision Transformer (ViT)~\cite{dosovitskiy2020VIT}) to establish a discriminative embedding space. Hyp-ViT employs a pre-trained ViT as a feature extractor, and learns image embeddings in the hyperbolic space. The empirical results verify the superiority of the Hyp-ViT configuration and endow a good practice of the hyperbolic space.

It shows that hyperbolic geometry brought significant performance gain for model adaptation learning in low data regimes, especially the FSL task. More advanced practice on other model adaptation learning paradigms, including continual learning, neural architecture search, dynamic neural network, \etc, on hyperbolic space also requires more efforts.

\begin{table}[!ht]
\begin{center}
\scalebox{0.77}
{
\begin{tabular}{llll}


\hline

\multirow{2}{*}{Ref.} & \multirow{2}{*}{Conf.} & \multirow{2}{*}{Appl.} & \multirow{2}{*}{Lear.}\\
&&&\\

\hline

\cite{Yoshihiro2019_WrappedND} & ICML & Image Generation & UL\\

\cline{3-4}

\cite{Nicholas_2019_KDD} & KDD & Clustering & UL\\

\cline{3-4}

\cite{Emile2019_HVAE} & NeurIPS & Image Generation & UL\\

\Xhline{3\arrayrulewidth}

\multirow{2}{*}{\cite{Khrulkov_2020_CVPR}} & \multirow{2}{*}{CVPR} & Few-shot Learning & \multirow{2}{*}{MAL}\\
&& Person Re-identification \\

\cline{3-4}

\cite{Liu_2020_CVPR} & CVPR & Zero-shot Learning & MAL\\

\cline{3-4}

\cite{Long_2020_CVPR} & CVPR & Action Search & SL\\

\cline{3-4}

\cite{Hyperbolic3D2020_ECCV} & ECCV & 3D Shape Recognition & SL\\

\cline{3-4}

\cite{Peng2020_ACMMM} & ACM MM & Action Recognition & SL\\

\cline{3-4}

\cite{NEURIPS2020_ac10ec1a_HypHC} & NeurIPS & Clustering & UL\\


\Xhline{3\arrayrulewidth}

\cite{suris2021hyperfuture} & CVPR & Action Prediction & SL\\

\cline{3-4}

\cite{Yan_2021_CVPR} & CVPR & Metric Learning & UL\\

\cline{3-4}

\cite{weng2021unsupervised_cvpr} & CVPR & Image Segmentation & UL\\

\cline{3-4}

\cite{Gao_2021_ICCV} & ICCV & Few-shot Learning & MAL\\

\cline{3-4}

&& Few-shot Learning\\
\multirow{2}{*}{\cite{Fang_2021_ICCV}} & \multirow{2}{*}{ICCV} & Zero-shot Learning & \multirow{2}{*}{MAL}\\
&&Person Re-identification\\
&&Knowledge Distillation\\

\cline{3-4}



\cite{NEURIPS2021_291d43c6} & NeurIPS & Image Segmentation & UL\\

\cline{3-4}

\cite{NEURIPS2021_01259a0c} & NeurIPS & Few-shot Learning & MAL\\

\Xhline{3\arrayrulewidth}

\cite{ma2022adaptive} & AAAI & Few-shot Learning & MAL\\

\cline{3-4}

\cite{Ola2022Fisheye} & AAAI & Image Recognition & SL\\

\cline{3-4}

\cite{Atigh_2022_CVPR} & CVPR & Image Segmentation & SL\\

\cline{3-4}

\multirow{2}{*}{\cite{Guo_2022_CVPR_Clipped}} & \multirow{2}{*}{CVPR} & Image Recognition & SL\\
&&Few-shot Learning & MAL\\

\cline{3-4}



\cite{Ermolov_2022_CVPR} & CVPR & Metric Learning & MAL\\

\cline{3-4}

\cite{CMHHC_IJCAI2022} & IJCAI & Clustering & UL\\

\cline{3-4}

\cite{HyperKT_IJCAI2022} & IJCAI & Few-shot Learning & MAL\\

\cline{3-4}

\cite{Yu2022_MICCAI} & MICCAI & Image Recognition & SL\\
\hline

\end{tabular}
}
\end{center}
\caption{Summary of the computer vision algorithms in hyperbolic space. In this table, ``Ref.'', ``Conf.'', ``Appl.'' and ``Lear.'' indicate ``Reference'', ``Conference'', ``Application'' and ``Learning Method'', respectively. ``UL'', ``SL'' and ``MAL'' are abbreviates for ``Unsupervised Learning'', ``Supervised Learning'' and ``Model Adaptation Learning'', respectively.}
\label{table:papers}
\end{table}

\section{Conclusion and Future Directions}\label{sec:conclusion}

In this paper, we review the advanced learning methods of hyperbolic geometry in CV applications. In contrast to the flat Euclidean space, hyperbolic geometry is a curved space, featured with negative curvature. This property enables the hyperbolic space to embed the hierarchical relations of the data, leading to powerful representation capacity. Recent studies vividly demonstrate how CV applications benefit from the hyperbolic space by exploring the hierarchical structure of the dataset. Also, a diversity of learning paradigms of CV applications, ranging from unsupervised learning, and supervised learning to model adaptation, are investigated in this manuscript to justify the superiority of hyperbolic geometry. We believe this brief survey will provide an overview of the developments of hyperbolic space in the CV domain, and encourages further improvements constrained by this geometric prior. In the following, we would like to discuss possible future directions of hyperbolic geometry in the CV field.

\textbf{Embedding space with mixed geometries.} In real-life practice, the underlying stricture of the data is complex. That said, embedding all visual data of a dataset in a fixed geometry is not an optimal solution. As suggested in \cite{GU_ICLR2019_Mixed-curvature,Skopek2020Mixed-curvature}, learning a hybrid embedding space, that includes both the flat space and curved space, could be a potential solution, and has been evaluated in the task of word embedding and graph reconstruction. In the CV community, learning geometry components and curvature values from visual data is still an open-problem and requires further studies. On top of leveraging a space with mixed curvatures, learning to adjust the curvature to data and problem is also a solution to explore the potential of the hyperbolic geometry.

\textbf{Advanced neural operators/algorithms in the hyperbolic space.} The recent extensive studies on hyperbolic geometry have extended a number of Euclidean neural components to the hyperbolic space, like auto-encoder, recurrent neural network, fully-connected layer \etc, for NLP or graph learning tasks~\cite{Wei2022_hyperbolic_PAMI}. In the CV domain, a hybrid architecture, discussed in \textsection~\ref{sec:SL}, is widely employed. To fully benefit from the hyperbolic geometry, it is necessary to explore developing feature extractor (\eg, CNNs and Transformer), network layer (\eg, attention mechanism) and learning paradigm (\eg, neural architecture search) in the hyperbolic space.

\textbf{Self-supervised hyperbolic geometry pre-training.} Neural network pre-training from large-scale datasets has shown its great potential in NLP and CV applications~\cite{bert_devlin2019,He_2022_CVPR}. The pre-trained models usually learn the knowledge from data in self-supervised learning (SSL), and can be seamlessly adapted to the downstream tasks via fine-tuning with few samples. This inspires us to build a pre-trained model that encodes the knowledge in the hyperbolic space. Intuitively, the tree structure exists in many natural scenarios, like biological taxonomy, or social networks. Thus a pre-trained hyperbolic space can learn such knowledge like human, enabling the intelligence machine to classify objects scientifically.

\textbf{Employing the benefits of other models of hyperbolic space in the CV applications.} Hyperbolic geometry can be modeled by five isometric models. In the CV domain, almost of approaches employ the Poincar\'e ball as the model of the hyperbolic space. While other AI applications, \eg, NLP or graph learning, benefit from the Lorentz model. This motivates us to develop methods that can choose the optimal model, which is data or problem-dependent.


\bibliographystyle{named}
\bibliography{ijcai23}

\begin{thebibliography}{}

\bibitem[\protect\citeauthoryear{Ahmad and Lecue}{2022}]{Ola2022Fisheye}
Ola Ahmad and Freddy Lecue.
\newblock Fisheyehdk: Hyperbolic deformable kernel learning for ultra-wide
  field-of-view image recognition.
\newblock In {\em Proceedings of the AAAI Conference on Artificial
  Intelligence}, volume~36, pages 5968--5975, 2022.

\bibitem[\protect\citeauthoryear{Akata \bgroup \em et al.\egroup
  }{2015}]{ALE_Akata_PAMI}
Zeynep Akata, Florent Perronnin, Zaid Harchaoui, and Cordelia Schmid.
\newblock Label-embedding for image classification.
\newblock {\em IEEE Transactions on Pattern Analysis and Machine Intelligence},
  pages 1425--1438, 2015.

\bibitem[\protect\citeauthoryear{Ambrogioni \bgroup \em et al.\egroup
  }{2018}]{NEURIPS2018_WVI}
Luca Ambrogioni, Umut G\"{u}\c{c}l\"{u}, Ya\u{g}mur G\"{u}\c{c}l\"{u}t\"{u}rk,
  Max Hinne, Marcel A.~J. van Gerven, and Eric Maris.
\newblock Wasserstein variational inference.
\newblock In {\em Advances in Neural Information Processing Systems},
  volume~31, 2018.

\bibitem[\protect\citeauthoryear{Atigh \bgroup \em et al.\egroup
  }{2021}]{NEURIPS2021_01259a0c}
Mina~Ghadimi Atigh, Martin Keller-Ressel, and Pascal Mettes.
\newblock Hyperbolic busemann learning with ideal prototypes.
\newblock In {\em Advances in Neural Information Processing Systems},
  volume~34, pages 103--115, 2021.

\bibitem[\protect\citeauthoryear{Atigh \bgroup \em et al.\egroup
  }{2022}]{Atigh_2022_CVPR}
Mina~Ghadimi Atigh, Julian Schoep, Erman Acar, Nanne van Noord, and Pascal
  Mettes.
\newblock Hyperbolic image segmentation.
\newblock In {\em Proceedings of the IEEE/CVF Conference on Computer Vision and
  Pattern Recognition}, pages 4453--4462, June 2022.

\bibitem[\protect\citeauthoryear{Cannon \bgroup \em et al.\egroup
  }{1997}]{hyperbolic_1997}
James~W. Cannon, William~J. Floyd, Richard Kenyon, and Walter~R. Parry.
\newblock {\em Hyperbolic Geometry}.
\newblock MSRI Publications, 1997.

\bibitem[\protect\citeauthoryear{Chami \bgroup \em et al.\egroup
  }{2019}]{Chami2019_HyperbolicGCNN_NIPS}
Ines Chami, Zhitao Ying, Christopher R{\'e}, and Jure Leskovec.
\newblock Hyperbolic graph convolutional neural networks.
\newblock In {\em Advances in Neural Information Processing Systems}, pages
  4869--4880, 2019.

\bibitem[\protect\citeauthoryear{Chami \bgroup \em et al.\egroup
  }{2020}]{NEURIPS2020_ac10ec1a_HypHC}
Ines Chami, Albert Gu, Vaggos Chatziafratis, and Christopher R\'{e}.
\newblock From trees to continuous embeddings and back: Hyperbolic hierarchical
  clustering.
\newblock In {\em Advances in Neural Information Processing Systems},
  volume~33, pages 15065--15076, 2020.

\bibitem[\protect\citeauthoryear{Chen \bgroup \em et al.\egroup
  }{2020}]{Hyperbolic3D2020_ECCV}
Jiaxin Chen, Jie Qin, Yuming Shen, Li~Liu, Fan Zhu, and Ling Shao.
\newblock Learning attentive and hierarchical representations for 3d shape
  recognition.
\newblock In {\em European Conference on Computer Vision}, pages 105--122,
  2020.

\bibitem[\protect\citeauthoryear{Dasgupta}{2016}]{Dasgupta_HC}
Sanjoy Dasgupta.
\newblock A cost function for similarity-based hierarchical clustering.
\newblock In {\em Proceedings of the 48th annual ACM symposium on Theory of
  Computing}, pages 118--127, 2016.

\bibitem[\protect\citeauthoryear{Davidson \bgroup \em et al.\egroup
  }{2018}]{SVAE18_UAI}
Tim~R. Davidson, Luca Falorsi, Nicola De~Cao, Thomas Kipf, and Jakub~M.
  Tomczak.
\newblock Hyperspherical variational auto-encoders.
\newblock In {\em 34th Conference on Uncertainty in Artificial Intelligence},
  2018.

\bibitem[\protect\citeauthoryear{Devlin \bgroup \em et al.\egroup
  }{2019}]{bert_devlin2019}
Jacob Devlin, Ming-Wei Chang, Kenton Lee, and Kristina Toutanova.
\newblock Bert: Pre-training of deep bidirectional transformers for language
  understanding.
\newblock In {\em 2019 Conference of the North American Chapter of the
  Association for Computational Linguistics: Human Language Technologies},
  pages 4171--4186, June 2019.

\bibitem[\protect\citeauthoryear{Dosovitskiy \bgroup \em et al.\egroup
  }{2021}]{dosovitskiy2020VIT}
Alexey Dosovitskiy, Lucas Beyer, Alexander Kolesnikov, Dirk Weissenborn,
  Xiaohua Zhai, Thomas Unterthiner, Mostafa Dehghani, Matthias Minderer, Georg
  Heigold, Sylvain Gelly, Jakob Uszkoreit, and Neil Houlsby.
\newblock An image is worth 16x16 words: Transformers for image recognition at
  scale.
\newblock In {\em International Conference on Learning Representations}, 2021.

\bibitem[\protect\citeauthoryear{Ermolov \bgroup \em et al.\egroup
  }{2022}]{Ermolov_2022_CVPR}
Aleksandr Ermolov, Leyla Mirvakhabova, Valentin Khrulkov, Nicu Sebe, and Ivan
  Oseledets.
\newblock Hyperbolic vision transformers: Combining improvements in metric
  learning.
\newblock In {\em Proceedings of the IEEE/CVF Conference on Computer Vision and
  Pattern Recognition}, pages 7409--7419, June 2022.

\bibitem[\protect\citeauthoryear{Fang \bgroup \em et al.\egroup
  }{2021}]{Fang_2021_ICCV}
Pengfei Fang, Mehrtash Harandi, and Lars Petersson.
\newblock Kernel methods in hyperbolic spaces.
\newblock In {\em Proceedings of the IEEE/CVF International Conference on
  Computer Vision}, pages 10665--10674, October 2021.

\bibitem[\protect\citeauthoryear{Ganea \bgroup \em et al.\egroup
  }{2018}]{Ganea2018_HyperbolicNN_NIPS}
Octavian Ganea, Gary B{\'e}cigneul, and Thomas Hofmann.
\newblock Hyperbolic neural networks.
\newblock In {\em Advances in neural information processing systems}, pages
  5345--5355, 2018.

\bibitem[\protect\citeauthoryear{Gao \bgroup \em et al.\egroup
  }{2021}]{Gao_2021_ICCV}
Zhi Gao, Yuwei Wu, Yunde Jia, and Mehrtash Harandi.
\newblock Curvature generation in curved spaces for few-shot learning.
\newblock In {\em Proceedings of the IEEE/CVF International Conference on
  Computer Vision}, pages 8691--8700, October 2021.

\bibitem[\protect\citeauthoryear{Gu \bgroup \em et al.\egroup
  }{2019}]{GU_ICLR2019_Mixed-curvature}
Albert Gu, Frederic Sala, Beliz Gunel, and Christopher Ré.
\newblock Learning mixed-curvature representations in product spaces.
\newblock In {\em International Conference on Learning Representations}, 2019.

\bibitem[\protect\citeauthoryear{Guo \bgroup \em et al.\egroup
  }{2022}]{Guo_2022_CVPR_Clipped}
Yunhui Guo, Xudong Wang, Yubei Chen, and Stella~X. Yu.
\newblock Clipped hyperbolic classifiers are super-hyperbolic classifiers.
\newblock In {\em Proceedings of the IEEE/CVF Conference on Computer Vision and
  Pattern Recognition}, pages 11--20, June 2022.

\bibitem[\protect\citeauthoryear{He \bgroup \em et al.\egroup
  }{2022}]{He_2022_CVPR}
Kaiming He, Xinlei Chen, Saining Xie, Yanghao Li, Piotr Doll\'ar, and Ross
  Girshick.
\newblock Masked autoencoders are scalable vision learners.
\newblock In {\em Proceedings of the IEEE/CVF Conference on Computer Vision and
  Pattern Recognition}, pages 16000--16009, June 2022.

\bibitem[\protect\citeauthoryear{Hinton \bgroup \em et al.\egroup
  }{2014}]{KD_hinton}
Geoffrey Hinton, Oriol Vinyals, and Jeff Dean.
\newblock Distilling the knowledge in a neural network.
\newblock In {\em Twenty-eighth Conference on Neural Information Processing
  Systems, Deep Learning Workshop}, 2014.

\bibitem[\protect\citeauthoryear{Ho \bgroup \em et al.\egroup
  }{2020}]{NIPS2020_DenoiseDiffusion}
Jonathan Ho, Ajay Jain, and Pieter Abbeel.
\newblock Denoising diffusion probabilistic models.
\newblock In {\em Advances in Neural Information Processing Systems},
  volume~33, pages 6840--6851, 2020.

\bibitem[\protect\citeauthoryear{Hsu \bgroup \em et al.\egroup
  }{2021}]{NEURIPS2021_291d43c6}
Joy Hsu, Jeffrey Gu, Gong Wu, Wah Chiu, and Serena Yeung.
\newblock Capturing implicit hierarchical structure in 3d biomedical images
  with self-supervised hyperbolic representations.
\newblock In {\em Advances in Neural Information Processing Systems},
  volume~34, pages 5112--5123, 2021.

\bibitem[\protect\citeauthoryear{Khrulkov \bgroup \em et al.\egroup
  }{2020}]{Khrulkov_2020_CVPR}
Valentin Khrulkov, Leyla Mirvakhabova, Evgeniya Ustinova, Ivan Oseledets, and
  Victor Lempitsky.
\newblock Hyperbolic image embeddings.
\newblock In {\em Proceedings of the IEEE/CVF Conference on Computer Vision and
  Pattern Recognition}, June 2020.

\bibitem[\protect\citeauthoryear{Kingma and Welling}{2014}]{Kingma2014_VAE}
Diederik~P Kingma and Max Welling.
\newblock Auto-encoding variational bayes.
\newblock In {\em International Conference on Learning Representations}, 2014.

\bibitem[\protect\citeauthoryear{Krishnamurthy \bgroup \em et al.\egroup
  }{2012}]{Akshay_HC_ICML}
Akshay Krishnamurthy, Sivaraman Balakrishnan, Min Xu, and Aarti Singh.
\newblock Efficient active algorithms for hierarchical clustering.
\newblock In {\em Proceedings of the 29th International Conference on Machine
  Learning}, pages 267--274, 2012.

\bibitem[\protect\citeauthoryear{Li and Hoiem}{2016}]{li2016learning}
Zhizhong Li and Derek Hoiem.
\newblock Learning without forgetting.
\newblock In {\em European Conference on Computer Vision}, pages 614--629,
  2016.

\bibitem[\protect\citeauthoryear{Li \bgroup \em et al.\egroup
  }{2014}]{LiWei2014DeepReID}
Wei Li, Rui Zhao, Tong Xiao, and Xiaogang Wang.
\newblock Deepreid: Deep filter paring neural network for person
  re-identification.
\newblock In {\em The IEEE Conference on Computer Vision and Pattern
  Recognition}, pages 1--8, June 2014.

\bibitem[\protect\citeauthoryear{Lin \bgroup \em et al.\egroup
  }{2022}]{CMHHC_IJCAI2022}
Fangfei Lin, Bing Bai, Yazhou Ren, Peng Zhao, and Zenglin Xu.
\newblock Contrastive multi-view hyperbolic hierarchical clustering.
\newblock In {\em 2022 International Joint Conference on Artificial
  Intelligence}, pages 3250--3256, 2022.

\bibitem[\protect\citeauthoryear{Liu \bgroup \em et al.\egroup
  }{2020}]{Liu_2020_CVPR}
Shaoteng Liu, Jingjing Chen, Liangming Pan, Chong-Wah Ngo, Tat-Seng Chua, and
  Yu-Gang Jiang.
\newblock Hyperbolic visual embedding learning for zero-shot recognition.
\newblock In {\em IEEE/CVF Conference on Computer Vision and Pattern
  Recognition}, June 2020.

\bibitem[\protect\citeauthoryear{Long \bgroup \em et al.\egroup
  }{2020}]{Long_2020_CVPR}
Teng Long, Pascal Mettes, Heng~Tao Shen, and Cees G.~M. Snoek.
\newblock Searching for actions on the hyperbole.
\newblock In {\em IEEE/CVF Conference on Computer Vision and Pattern
  Recognition}, June 2020.

\bibitem[\protect\citeauthoryear{Ma \bgroup \em et al.\egroup
  }{2022}]{ma2022adaptive}
Rongkai Ma, Pengfei Fang, Tom Drummond, and Mehrtash Harandi.
\newblock Adaptive poincar{\'e} point to set distance for few-shot
  classification.
\newblock In {\em Proceedings of the AAAI Conference on Artificial
  Intelligence}, volume~36, pages 1926--1934, 2022.

\bibitem[\protect\citeauthoryear{Mathieu \bgroup \em et al.\egroup
  }{2019}]{Emile2019_HVAE}
Emile Mathieu, Charline Le~Lan, Chris~J. Maddison, Ryota Tomioka, and Yee~Whye
  Teh.
\newblock Continuous hierarchical representations with poincar\'{e} variational
  auto-encoders.
\newblock In {\em Advances in Neural Information Processing Systems},
  volume~32, pages 12565--12576, 2019.

\bibitem[\protect\citeauthoryear{Monath \bgroup \em et al.\egroup
  }{2019}]{Nicholas_2019_KDD}
Nicholas Monath, Manzil Zaheer, Daniel Silva, Andrew McCallum, and Amr Ahmed.
\newblock Gradient-based hierarchical clustering using continuous
  representations of trees in hyperbolic space.
\newblock In {\em 25th ACM SIGKDD International Conference on Knowledge
  Discovery and Data Mining}, July 2019.

\bibitem[\protect\citeauthoryear{Murtagh and
  Contreras}{2011}]{Flonn_hierarchical_clustering}
Flonn Murtagh and Pedro Contreras.
\newblock Algorithms for hierarchical clustering: an overview.
\newblock {\em WIREs Data Mining and Knowledge Discovery}, pages 86--97, 2011.

\bibitem[\protect\citeauthoryear{Nagano \bgroup \em et al.\egroup
  }{2019}]{Yoshihiro2019_WrappedND}
Yoshihiro Nagano, Shoichiro Yamaguchi, Yasuhiro Fujita, and Masanori Koyama.
\newblock A wrapped normal distribution on hyperbolic space for gradient-based
  learning.
\newblock In {\em The 36th International Conference on Machine Learning}, 2019.

\bibitem[\protect\citeauthoryear{Peng \bgroup \em et al.\egroup
  }{2020}]{Peng2020_ACMMM}
Wei Peng, Jingang Shi, Zhaoqiang Xia, and Guoying Zhao.
\newblock Mix dimension in poincar{\'e} geometry for 3d skeleton-based action
  recognition.
\newblock In {\em Proceedings of the 28th ACM International Conference on
  Multimedia}, 2020.

\bibitem[\protect\citeauthoryear{Peng \bgroup \em et al.\egroup
  }{2021}]{Wei2022_hyperbolic_PAMI}
Wei Peng, Tuomas Varanka, Abdelrahman Mostafa, Henglin Shi, and Guoying Zhao.
\newblock Hyperbolic deep neural networks: A survey.
\newblock {\em IEEE Transactions on Pattern Analysis and Machine Intelligence},
  2021.

\bibitem[\protect\citeauthoryear{Rezende \bgroup \em et al.\egroup
  }{2014}]{pmlr-v32-rezende14}
Danilo~Jimenez Rezende, Shakir Mohamed, and Daan Wierstra.
\newblock Stochastic backpropagation and approximate inference in deep
  generative models.
\newblock In {\em Proceedings of the 31st International Conference on Machine
  Learning}, volume~32, pages 1278--1286, 2014.

\bibitem[\protect\citeauthoryear{Schroff \bgroup \em et al.\egroup
  }{2015}]{Facenet_CVPR15}
Florian Schroff, Dmitry Kalenichenko, and James Philbin.
\newblock Facenet: A unified embedding for face recognition and clustering.
\newblock In {\em 2015 IEEE Conference on Computer Vision and Pattern
  Recognition}, pages 815--823, 2015.

\bibitem[\protect\citeauthoryear{Skopek \bgroup \em et al.\egroup
  }{2020}]{Skopek2020Mixed-curvature}
Ondrej Skopek, Octavian-Eugen Ganea, and Gary Bécigneul.
\newblock Mixed-curvature variational autoencoders.
\newblock In {\em International Conference on Learning Representations}, 2020.

\bibitem[\protect\citeauthoryear{Snell \bgroup \em et al.\egroup
  }{2017}]{ProtoNet_NIPS}
Jake Snell, Kevin Swersky, and Richard Zemel.
\newblock Prototypical networks for few-shot learning.
\newblock In {\em The Thirty-first Annual Conference on Neural Information
  Processing Systems}, pages 4077--4087, 2017.

\bibitem[\protect\citeauthoryear{Sur\'is \bgroup \em et al.\egroup
  }{2021}]{suris2021hyperfuture}
D\'idac Sur\'is, Ruoshi Liu, and Carl Vondrick.
\newblock Learning the predictability of the future.
\newblock In {\em Proceedings of the IEEE/CVF Conference on Computer Vision and
  Pattern Recognition}, pages 12607--12617, June 2021.

\bibitem[\protect\citeauthoryear{Tran \bgroup \em et al.\egroup
  }{2020}]{VinhTran2020HyperMLAB}
Lucas~Vinh Tran, Yi~Tay, Shuai Zhang, G.~Cong, and Xiaoli Li.
\newblock Hyperml: A boosting metric learning approach in hyperbolic space for
  recommender systems.
\newblock In {\em Proceedings of the 13th International Conference on Web
  Search and Data Mining}, 2020.

\bibitem[\protect\citeauthoryear{van~den Oord \bgroup \em et al.\egroup
  }{2017}]{NIPS2017_VQVAE}
Aaron van~den Oord, Oriol Vinyals, and koray kavukcuoglu.
\newblock Neural discrete representation learning.
\newblock In {\em Advances in Neural Information Processing Systems},
  volume~30, pages 6306--6315, 2017.

\bibitem[\protect\citeauthoryear{Weng \bgroup \em et al.\egroup
  }{2021}]{weng2021unsupervised_cvpr}
Zhenzhen Weng, Mehmet~Giray Ogut, Shai Limonchik, and Serena Yeung.
\newblock Unsupervised discovery of the long-tail in instance segmentation
  using hierarchical self-supervision.
\newblock In {\em Proceedings of the IEEE/CVF Conference on Computer Vision and
  Pattern Recognition}, pages 2603--2612, 2021.

\bibitem[\protect\citeauthoryear{Xu and Tian}{2015}]{Dongkuan_clustering}
Dongkuan Xu and Yingjie Tian.
\newblock A comprehensive survey of clustering algorithms.
\newblock {\em Annals of Data Science}, pages 165--193, 2015.

\bibitem[\protect\citeauthoryear{Yan \bgroup \em et al.\egroup
  }{2021}]{Yan_2021_CVPR}
Jiexi Yan, Lei Luo, Cheng Deng, and Heng Huang.
\newblock Unsupervised hyperbolic metric learning.
\newblock In {\em Proceedings of the IEEE/CVF Conference on Computer Vision and
  Pattern Recognition}, pages 12465--12474, June 2021.

\bibitem[\protect\citeauthoryear{Yu \bgroup \em et al.\egroup
  }{2018}]{Yu2018SlimmableNN}
Jiahui Yu, Linjie Yang, Ning Xu, Jianchao Yang, and Thomas Huang.
\newblock Slimmable neural networks.
\newblock In {\em Sixth International Conference on Learning Representations},
  pages 1--12, 2018.

\bibitem[\protect\citeauthoryear{Yu \bgroup \em et al.\egroup
  }{2022}]{Yu2022_MICCAI}
Zhen Yu, Toan Nguyen, Yaniv Gal, Lie Ju, Shekhar~S. Chandra, Lei Zhang, Paul
  Bonnington, Victoria Mar, Zhiyong Wang, and Zongyuan Ge.
\newblock Skin lesion recognition with class-hierarchy regularized hyperbolic
  embeddings.
\newblock In {\em International Conference on Medical Image Computing and
  Computer-Assisted Intervention}, pages 594--603, 2022.

\bibitem[\protect\citeauthoryear{Zhang \bgroup \em et al.\egroup
  }{2022}]{HyperKT_IJCAI2022}
Baoquan Zhang, Hao Jiang, Shanshan Feng, Xutao Li, Yunming Ye, and Rui Ye.
\newblock Hyperbolic knowledge transfer with class hierarchy for few-shot
  learning.
\newblock In {\em 2022 International Joint Conference on Artificial
  Intelligence}, pages 3723--3729, 2022.

\end{thebibliography}

\end{document}